\documentclass[conference]{IEEEtran}
\IEEEoverridecommandlockouts
\usepackage{cite}
\usepackage{algorithmic}
\usepackage{textcomp}

\usepackage[utf8]{inputenc}
\usepackage[T1]{fontenc}
\usepackage[english]{babel}
\usepackage{epsfig}
\usepackage{graphicx}
\usepackage{calc}
\usepackage{amssymb}
\usepackage{amstext}
\usepackage{amsmath}
\usepackage{mathtools}
\usepackage{amsthm}
\usepackage{multicol}
\usepackage{hyperref}
\usepackage{smartdiagram}
\usepackage{booktabs}
\usepackage{lmodern}
\usepackage{nicefrac}
\usepackage{siunitx}
\usepackage{caption}
\usepackage{subcaption}
\usepackage{rotating}

\newsavebox\CBox
\def\textBF#1{\sbox\CBox{#1}\resizebox{\wd\CBox}{\ht\CBox}{\textbf{#1}}}

\begin{document}

\title{Cubes3D: Neural Network based Optical Flow in Omnidirectional Image Scenes
\thanks{This work is funded by  the  European Regional  Development  Fund  (ERDF) and the Free State of Saxony under the grant number 100-241-945.}
}

\author{\IEEEauthorblockN{1\textsuperscript{st} André Apitzsch}
\IEEEauthorblockA{\textit{Dept. of Information Technology} \\
\textit{Chemnitz University of Technology}\\
Chemnitz, Germany \\
andre.apitzsch@etit.tu-chemnitz.de}
\and
\IEEEauthorblockN{2\textsuperscript{nd} Roman Seidel}
\IEEEauthorblockA{\textit{Dept. of Information Technology} \\
\textit{Chemnitz University of Technology}\\
Chemnitz, Germany \\
roman.seidel@etit.tu-chemnitz.de}
\and
\IEEEauthorblockN{3\textsuperscript{rd} Gangolf Hirtz}
\IEEEauthorblockA{\textit{Dept. of Information Technology} \\
\textit{Chemnitz University of Technology}\\
Chemnitz, Germany \\
g.hirtz@etit.tu-chemnitz.de}
}

\maketitle

\begin{abstract}
Optical flow estimation with convolutional neural networks (CNNs) has recently solved various tasks of computer vision successfully.
In this paper we adapt a state-of-the-art approach for optical flow estimation to omnidirectional images.
We investigate CNN architectures to determine high motion variations caused by the geometry of fish-eye images.
Further we determine the qualitative influence of texture on the non-rigid object to the motion vectors.
For evaluation of the results we create ground truth motion fields synthetically.
The ground truth contains cubes with static background. We test variations of pre-trained FlowNet~2.0 architectures by indicating common error metrics.
We generate competitive results for the motion of the foreground with inhomogeneous texture on the moving object.
\end{abstract}


\section{Introduction}
\label{sec:introduction}
Even for large displacements the estimation of optical flow is still a challenge in computer vision.
Common benchmark datasets like Middlebury Flow \cite{Baker2011},
KITTI Flow \cite{Menze2015CVPR} or Sintel \cite{Butler2012} address the environmental conditions in appearance change,
large motions and homogeneous image regions.
In this paper we investigate two central aspects: first, the generation of ground truth for optical flow.  
Second, the exploration of different architectures of correlating images using convolutional neural networks.
As a result, we apply various combinations of architectures of FlowNet~2.0 \cite{IMKDB17} to estimate the pixel displacements depending
on the regions of an image with omnidirectional camera geometry.
A typical application for obtaining optical flow from image sequences is motion segmentation in indoor scenes with a fixed camera position.

The remainder of this paper is structured as follows:
Section \ref{sec:related_work} presents previous research activities in estimation of the optical flow using deep neural networks.
Section \ref{sec:flow_omni} illustrates the applied network architectures of the neural network as well as the characteristics of images from a fish-eye camera.
Section \ref{sec:gt} explains how our ground truth is generated.
Section \ref{sec:experim_res} describes our experiments for computation of the optical flow and compares the results to our generated synthetic ground truth based on common error metrics.
Section \ref{sec:conclu} summarizes the paper's content, concludes our observations and gives ideas for future work.
The source code of our work is available at \url{https://gitlab.com/auxilia/cubes3d}.

\section{Related Work}
\label{sec:related_work}
While optical flow estimation on images from omnidirectional cameras is not well explored, a phase based method for optical flow estimation on perspective image data was provided by \cite{Alibouch2012}.
Reference \cite{Gautama2002} applied spatial filters (here: Gabor filters) to estimate valid component velocities with a recurrent neural network through the direct measurement of the phase linearity,
which is a good indicator of the motion velocity.
Beside the variational methods to compute optical flow from \cite{Horn1980} which produce a dense flow field,
the work of \cite{ChenK16} concentrates on the optimization of the variational objective.
The motivation for using global optimization methods is the limitation of the variational objective to small places to compute optical flow for large pixel displacements.
Adapting this approach to omnidirectional images helps to determine the correct motion fields.
\setlength{\fboxsep}{0pt}
\begin{figure}[t!]
    \centering
    \begin{subfigure}[t]{.32\linewidth}
        \fbox{\includegraphics[width=\textwidth]{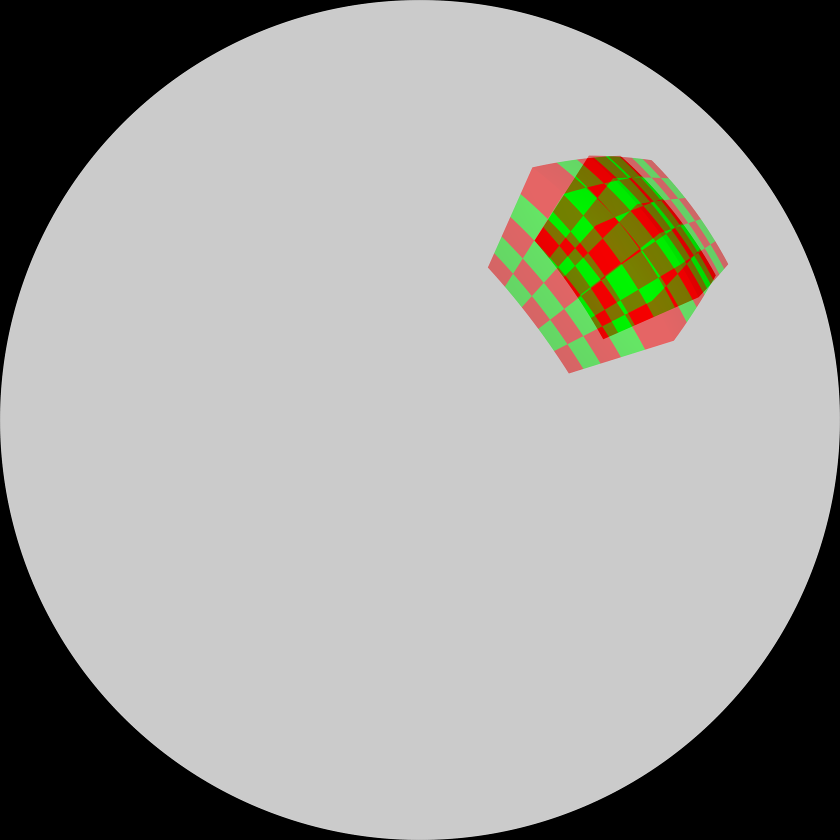}}
        \caption{input (overlay)}\label{fig:raw_data_omni}
    \end{subfigure}
    \begin{subfigure}[t]{.32\linewidth}
        \fbox{\includegraphics[width=\textwidth]{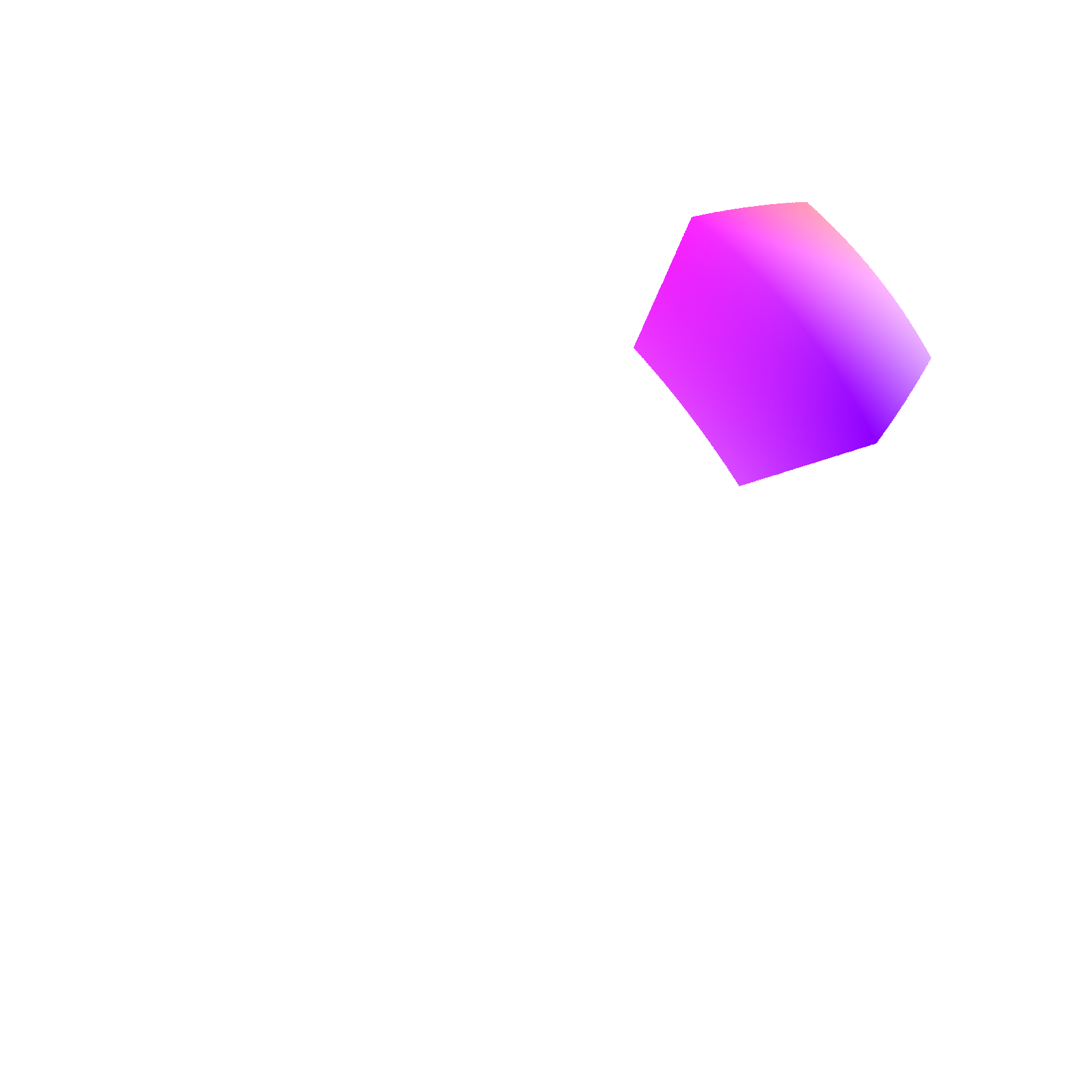}}
        \caption{ground truth}\label{fig:groundtruth}
    \end{subfigure}
    \begin{subfigure}[t]{.32\linewidth}
        \fbox{\includegraphics[width=\textwidth]{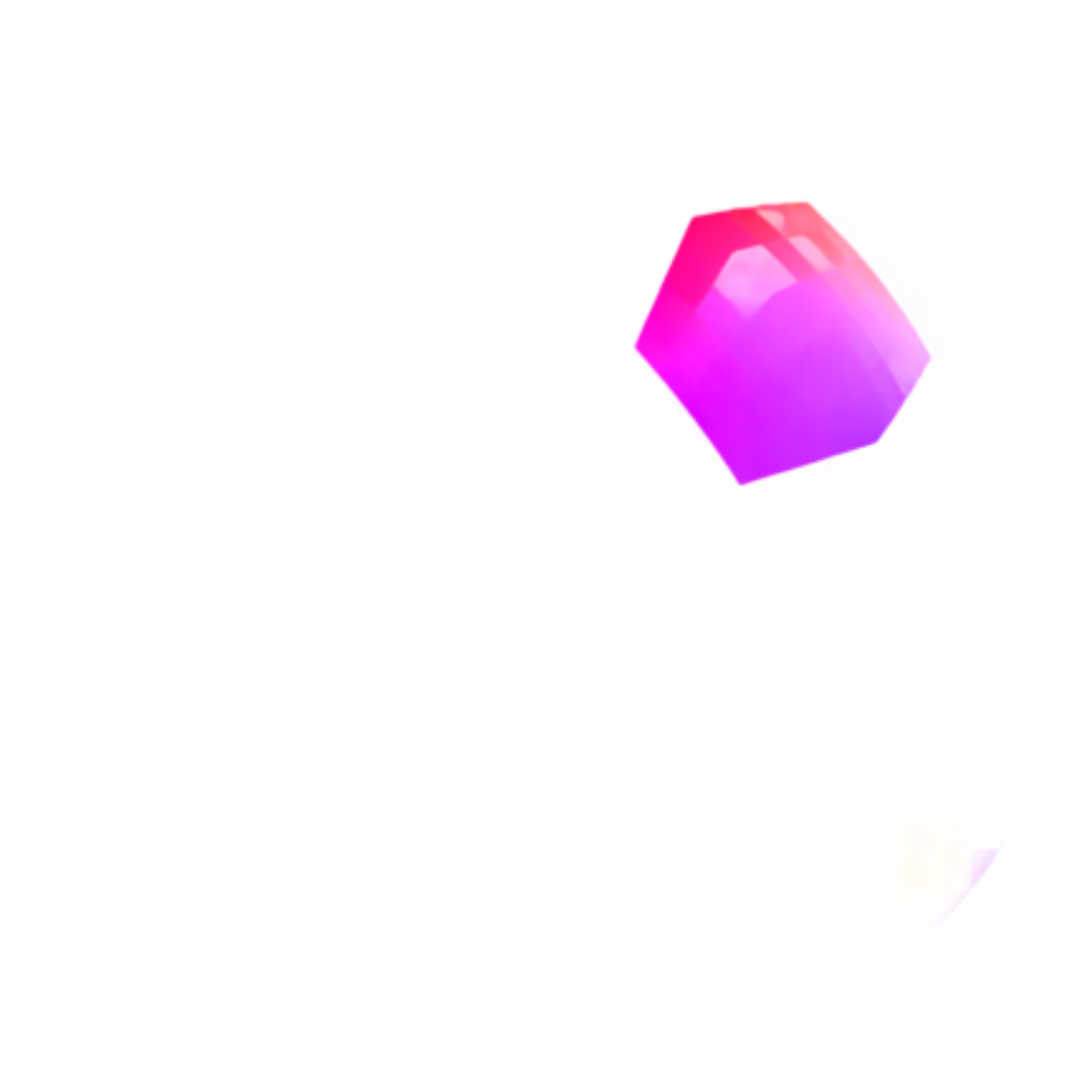}}
        \caption{estimation}\label{fig:flownet2_css}
    \end{subfigure}
    \caption{Optical flow of a moving 3-D cube in an omnidirectional image. Two consecutive frames of omnidirectional images (\ref{fig:raw_data_omni}), own generated ground truth (\ref{fig:groundtruth}) and a result of motion estimation with FlowNet2-CSS-ft-sd (\ref{fig:flownet2_css})(color-coding see \ref{fig:colorwheel}).}\label{fig:of_omni}
\end{figure}

In \cite{TranBFTP15} a deep neural network with 3-D convolutions was trained end-to-end to perform prediction at voxel-level.
The architecture leads to competitive results in semantic segmentation, video coloring and optical flow estimation.

As an alternative approach, which we follow, the optical flow was interpreted as a learning problem called FlowNet \cite{DFIB15} and Flownet~2.0 \cite{IMKDB17} based on neural networks.
The considerable improvements in quality and speed are caused by three major aspects:
first, accurate training datasets (e.g. FlyingThings3D \cite{MIFDB16} and FlyingChairs \cite{DFIB15}) are responsible for the success of supervised learning.
Second, the stacked network architecture with image warping of the second image with intermediate optical flow. 
Third, the consideration of small pixel displacements by introducing a subnetwork. 


\section{Optical Flow on Omnidirectional Images}
\label{sec:flow_omni}

\subsection{Omnidirectional Camera Model}

The camera model describes the transformation of a 3-D scene point to the coordinates of a 2-D image point.
We use the omnidirectional camera model with an equidistant projection, described in \cite{KannalaB06}.
In general the radial projection of an incoming light ray \(\mathbf{\Phi} = (\theta, \varphi)^T\) onto a virtual image plane with a distance of \(f = 1\) to the projection center can be modelled with the radial projection function:
\begin{equation}
    \mathbf{X_{norm}} = \mathcal{F}\left(\mathbf{\Phi}\right) = \rho(\theta) \mathbf{u}_r(\varphi)
\label{eq:proj_general}
\end{equation}
where $\rho(\theta)$ is the radius of the projected image point with respect to the center of the image,
\(\theta\) is the angle between the incoming ray and the optical axis and \(\mathbf{u}_r(\varphi) = (\cos{\varphi}, \sin{\varphi})^T\) is
the unit vector in radial direction.
The projection of the pinhole camera model has a singularity for \(\theta = \nicefrac{\pi}{2}\) leading to a limitation of a field of view smaller than \ang{180}.
Fish-eye lenses are designed to cover the full hemisphere in front of the camera.
One of the projection functions is the \textit{equidistant projection}:
\begin{equation}
    \rho(\theta) = \theta
\label{eq:proj_linear}
\end{equation}
where the distance of the mapped image point to the center of the image is linearly proportional to the angle \(\theta\) of the incoming ray.

\subsection{Convolutional Neural Networks for Optical Flow}

\begin{figure*}[t!]
\centering
    \includegraphics[width=\textwidth]{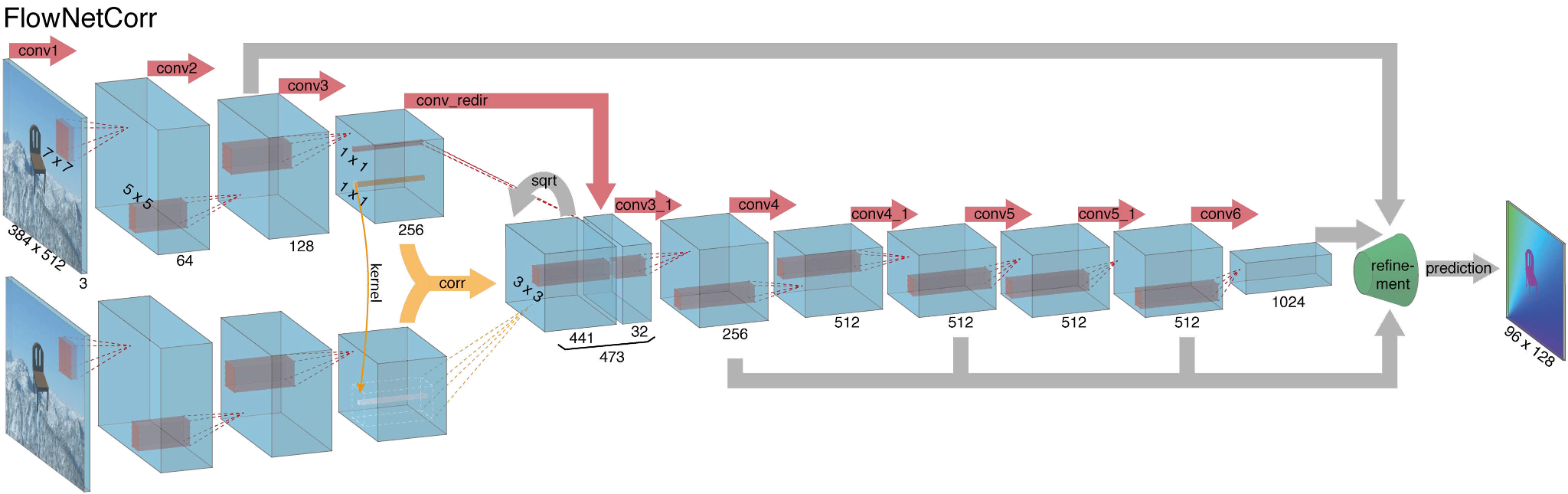}
\caption{Correlated FlowNet Architecture (FlowNetCorr) by \cite{DFIB15}.
Creating two parallel processing streams to correlate the feature-maps on pixel level and combine them on a higher level.
Finding correspondences is realized through a $\textit{correlation layer}$ by comparing patches of two feature maps.}
\label{fig:flownetcorr}
\end{figure*}

Computing optical flow in 2-D image space is the estimation of the motion vector of two consecutive frames.
The idea is to learn input-output relations from a sufficient amount of labeled data with a convolutional neural
network in the proposed FlowNet \cite{DFIB15} and FlowNet~2.0 \cite{IMKDB17}, respectively.
The general working principle of the network architecture of FlowNet2, based on FlowNetCorr is shown in Fig.~\ref{fig:flownetcorr}.

Finding correspondences by using correlation layer determines the differences between the resulted feature maps.
The approach is similar to convolving an image patch with the difference to convolve the first with the second patch of the feature map.
To consider the easiest step for two patches with a square patch of size \(K := 2k+1 \), the correlation of two patches at \(\mathbf{x_{1}}\) in the first map and \(\mathbf{x_{2}}\) in the second feature map is shown as:
\begin{equation}
 c(\mathbf{x_{1}}, \mathbf{x_{2}}) = \sum_{\mathclap{\substack{\mathbf{o}\in \lbrack -k,k \rbrack \times \lbrack -k,k \rbrack}}} \langle \mathbf{f_{1}(x_{1}+o), f_{2}(x_{2}+o)}\rangle
 \label{equ:correlation_cnn}
\end{equation}
where $\mathbf{f_{1}}$, $\mathbf{f_{2}}$ are the multi-channel feature maps. 
Instead of the convolution of an image with a filter, \eqref{equ:correlation_cnn} describes the convolution between two feature maps.

\begin{figure}
\centering
    \includegraphics[width=0.45\textwidth]{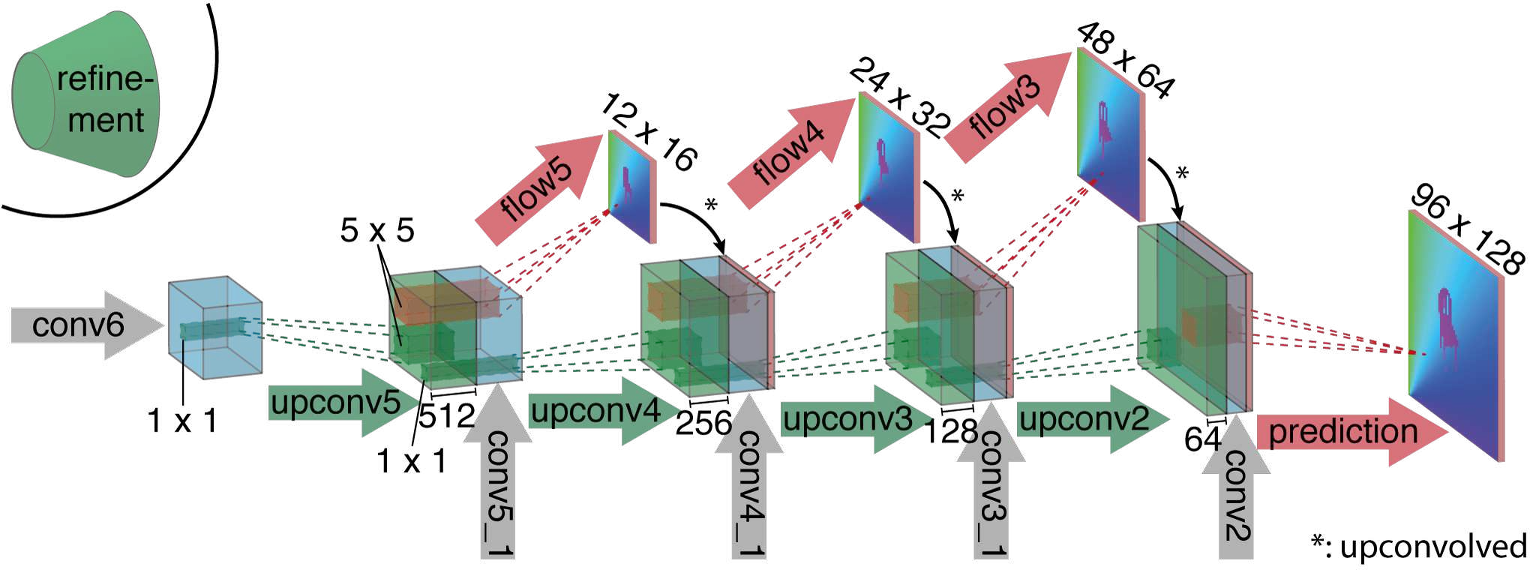}
\caption{In order to avoid pooling layers leading to a reduced resolution of the feature maps, a refinement of FlowNet architecture was done by \cite{DFIB15}.
The idea behind the refinement is to replace the pooling step with upconvolutional layers, as previously presented in \cite{DB15,DBLP:LongSD14,zeiler2011_deconv}.
The refinement is done with the help of a priori information from the previous layer with coarse optical flow, if available.}
\label{fig:refinement_flownetcorr}
\end{figure}

The refinement step of the architecture is shown in Fig.~\ref{fig:refinement_flownetcorr}, which is part of the FlowNet2-CSS-ft-sd architecture.
The range of the movement is limited to avoid the computation of all patch combinations for efficient forward and backward passes.

One of the main differences between FlowNet and FlowNet~2.0 is the idea of stacking networks for different quantity of motions in combination
with warping the second image.

\subsection{Architectures of FlowNet~2.0}
\label{architectures}

As our starting point we apply three different architectures of a deep neural network, namely FlowNet~2.0.
The original FlowNet2, the FlowNet2-SD and the FlowNet2-CSS-ft-sd.
While FlowNet2 leads to good results at flow on Sintel and Middlebury (average angular error on test) and a high average precision at Motion Segmentation, the FlowNet2-SD output comes up with smoother
results with less noise at smaller displacements even without refinement \cite{IMKDB17}.
We choose these architectures for the following reasons:
\paragraph{FlowNet2}

We pick up the idea of stacked networks of the FlowNet2 network architecture to handle the pronounced high variations of displacement in omnidirectional images.

\paragraph{FlowNet2-SD}

To handle small displacements, which are common by increased radius from the center of the fish-eye image, we compare the results of FlowNet2-SD
with our ground truth.

\paragraph{FlowNet2-CSS-ft-sd}

Here we consider the idea of training a network on a combination of Things3D and ChairsSDHom (Chairs Small Displacements Homogeneous, see \cite{IMKDB17}).
The architecture decreases the impact of large displacements by a non-linearity of the endpoint error compared to the FlowNet2-SD architecture.

\subsection{Optical Flow Color Coding}
\begin{figure}
\centering
    \includegraphics[width=0.15\textwidth]{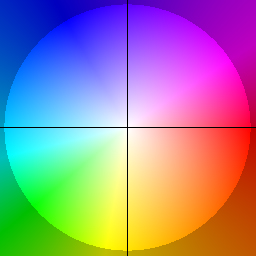}
\caption{Flow field coding used in this paper, following \cite{Butler2012}. The color indicates the direction of the motion vector.
The saturation enhances with increasing radius, which leads to strong movement at fields with high saturation.}
\label{fig:colorwheel}
\end{figure}

For visualization of the computed flow field, we use the color wheel of \cite{Butler2012}, which is shown in Fig.~\ref{fig:colorwheel}.

\section{Ground Truth Data}
\label{sec:gt}

Creating a real-world database for pixel-accurate optical flow ground truth is difficult \cite{Baker2011}.
To control various conditions in obtaining ground truth, namely scene illumination, pixel accuracy, non-rigid movements of the scene
we generate synthetic ground truth data as shown in Fig.~\ref{fig:groundtruth} using Blender.
The synthetic data are image sequences of a moving cube with homogeneous and inhomogeneous colored surfaces captured at 24 frames per second
by a virtual omnidirectional camera with equidistant projection (see section \ref{sec:flow_omni}).
The camera's fish-eye lens has a field of view of \ang{180} and an azimuth of \ang{360}.
For the ground truth flow we use the speed vector Blender calculates for each shader data. 
The cube moves along three different paths while staying in a fixed z-plane which is parallel to the image plane.
One path goes from left to right through the center of the image (\textit{linec}),
another one parallel to the first one but shifted to a different \(y\) plane (\textit{line});
and the last path is a spiral \textit{non-uniform rational basis spline} (NURBS)\cite{Rogers2000} curve that starts in the center and ends at the border of the image (\textit{spiral}).
For each path we record three ground truth sequences with different uniform velocities (1, 2 and 4 unit of speed).

\section{Experimental Results}
\label{sec:experim_res}

To capture image data we use a fixed camera with known intrinsic and extrinsic calibration \cite{mfin2014}.
The sensor was mounted on a height of \(z=\SI{2.5}{\meter}\) and the cube has the dimension of \(\SI{2}{\meter} \times \SI{2}{\meter}\times \SI{2}{\meter}\).
The center of the cube is always at \(z=\SI{0}{\meter}\).

We compare the results of three network architectures of FlowNet~2.0 to our own generated ground truth.
We pick up the idea of stacking networks from FlowNet~2.0.
Estimating the optical flow on images from omnidirectional camera geometry means to deal with peculiarity of different quantity of motions depending on the location of the object in the image.
The closer the object is located to the center of the image \((\nicefrac{x}{2}, \nicefrac{y}{2})\) the bigger the apparent movement.

\subsection{Qualitative Results}

\begin{figure*}[ht!]
\centering
    \begin{sideways}
        \small spiral-4, Frames 30\textrightarrow 31
    \end{sideways}
    \begin{subfigure}[b]{.19\linewidth}
        \caption{input (overlay)}\label{fig:input}
        \fbox{\includegraphics[width=\textwidth]{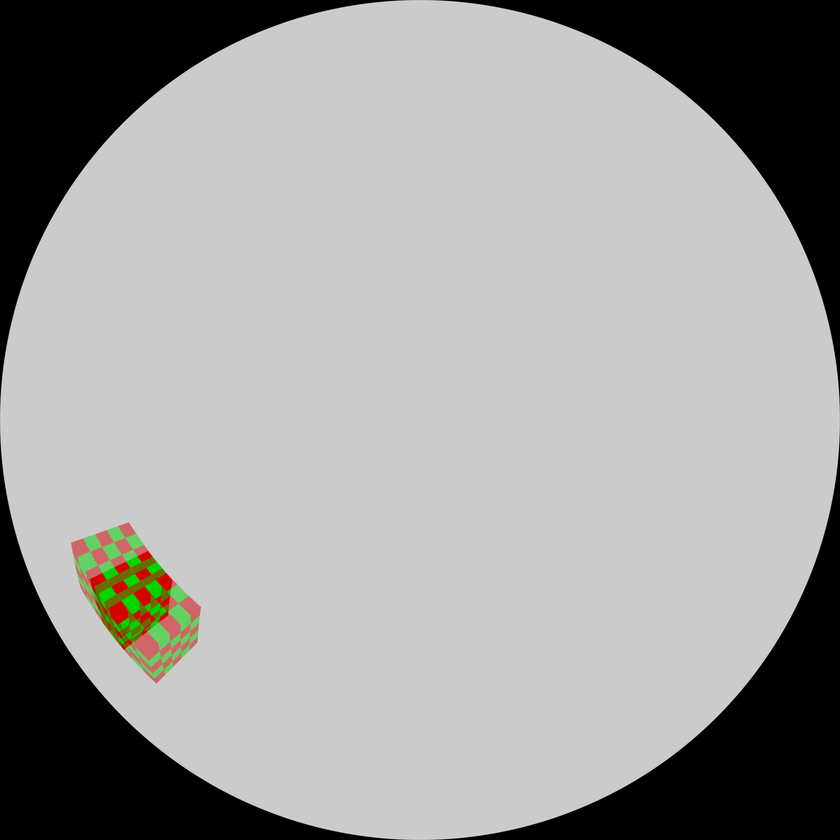}}
    \end{subfigure}
    \begin{subfigure}[b]{.19\linewidth}
        \caption{Ground truth}\label{fig:Groundtruth}
        \fbox{\includegraphics[width=\textwidth]{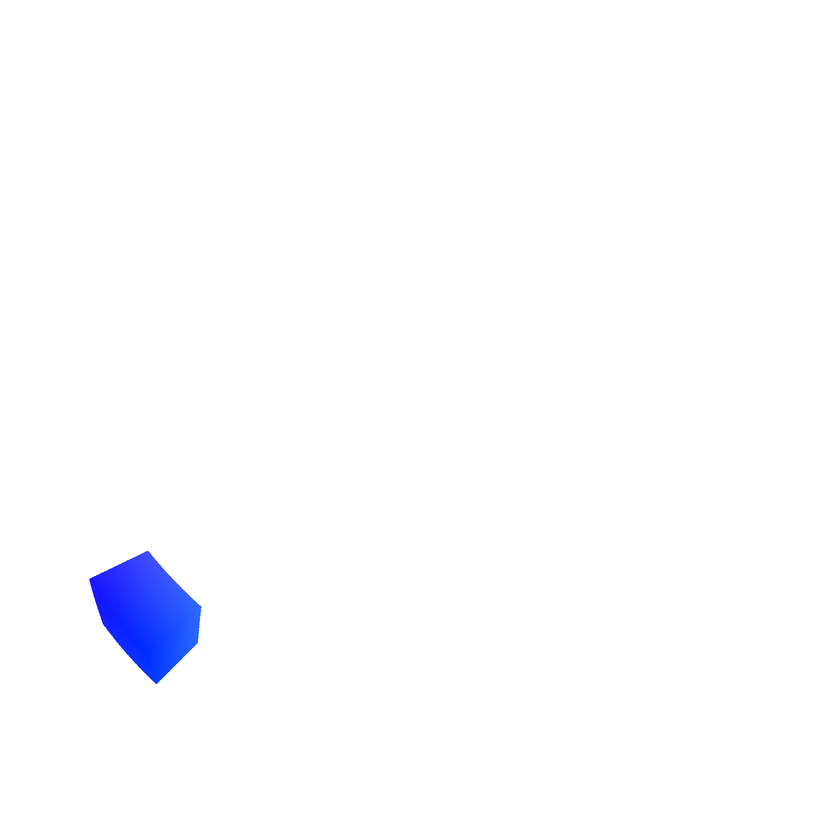}}
    \end{subfigure}
    \begin{subfigure}[b]{.19\linewidth}
        \caption{FlowNet2}\label{fig:FlowNet2}
        \fbox{\includegraphics[width=\textwidth]{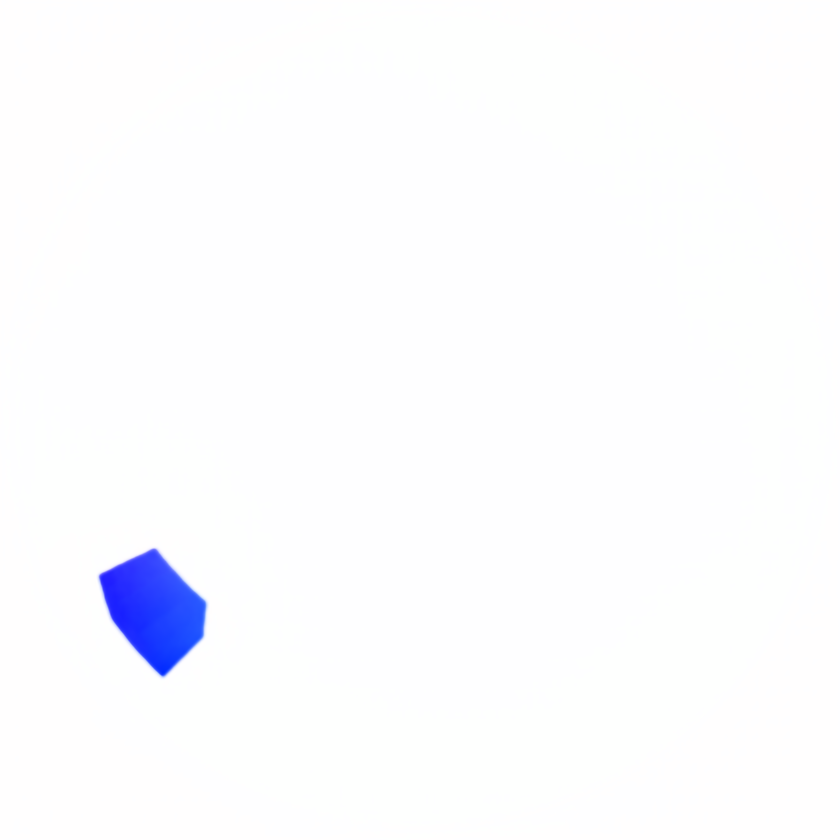}}
    \end{subfigure}
    \begin{subfigure}[b]{.19\linewidth}
        \caption{FN2-SD}\label{fig:FN2-SD}
        \fbox{\includegraphics[width=\textwidth]{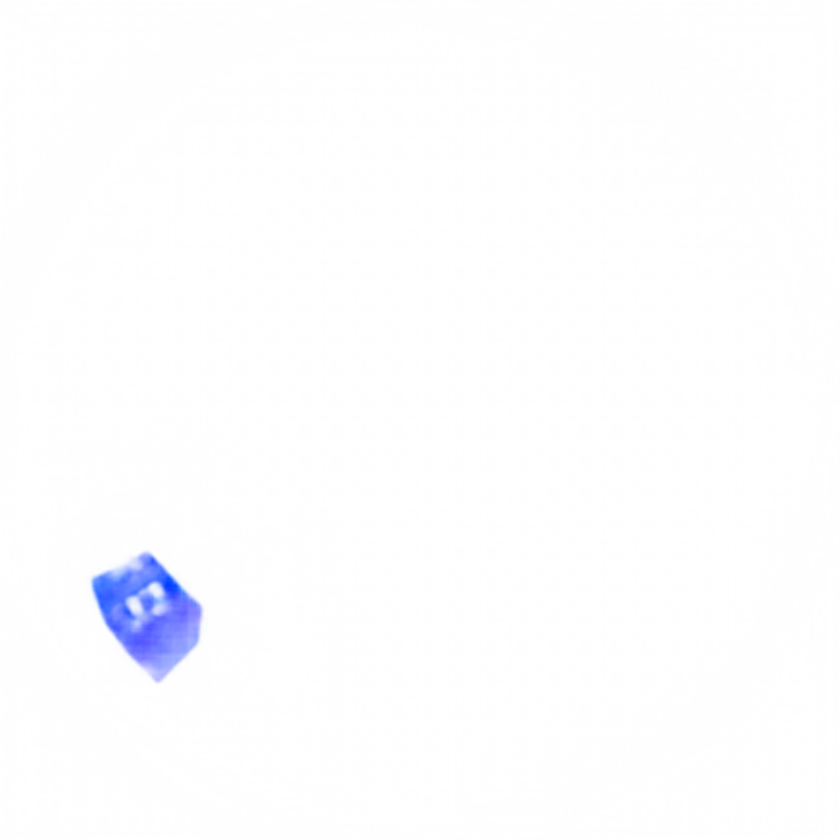}}
    \end{subfigure}
    \begin{subfigure}[b]{.19\linewidth}
        \caption{FN2-CSS-ft-sd}\label{fig:FN2-CSS-ft-sd}
        \fbox{\includegraphics[width=\textwidth]{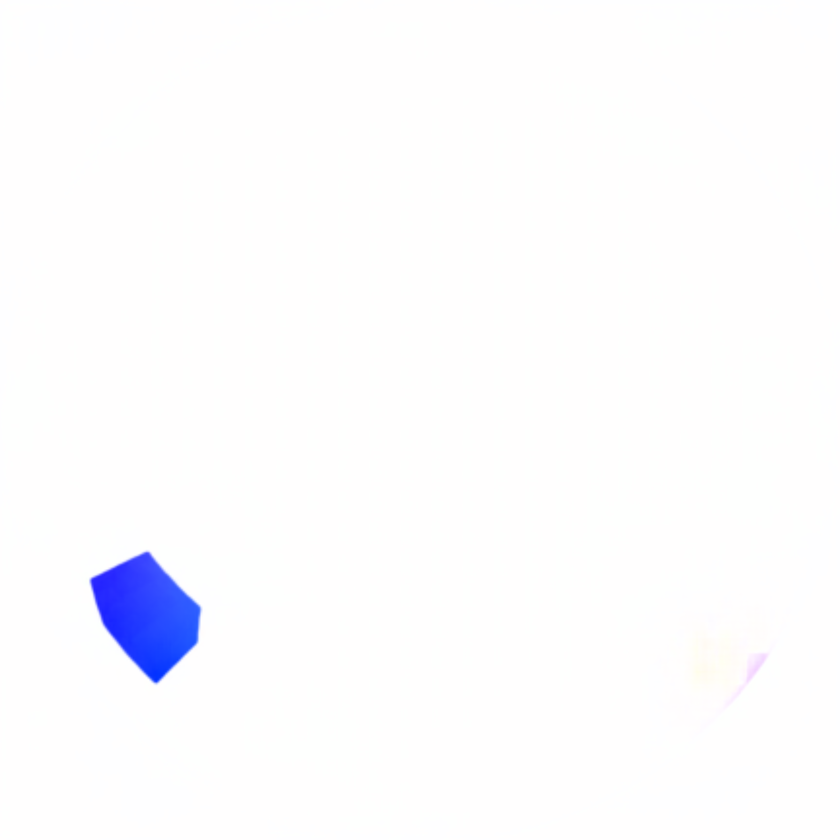}}
    \end{subfigure}

    \begin{sideways}
        \small spiral-1, Frames 53\textrightarrow 54
    \end{sideways}
    \begin{subfigure}[b][0.195\textwidth]{.19\linewidth}\fbox{\includegraphics[width=\textwidth]{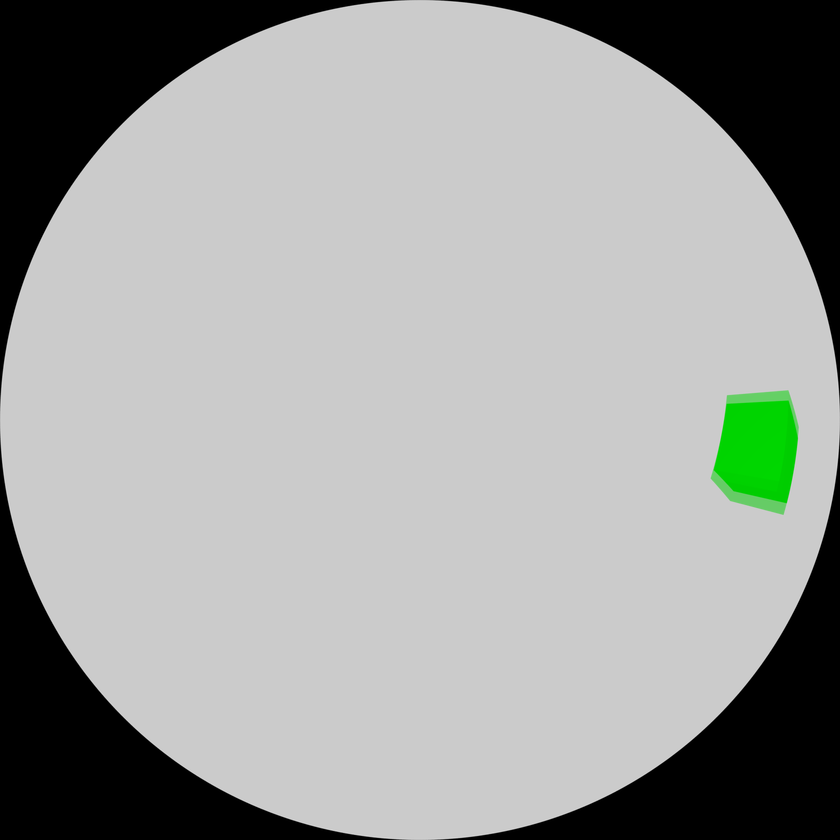}}\end{subfigure}
    \begin{subfigure}[b]{.19\linewidth}\fbox{\includegraphics[width=\textwidth]{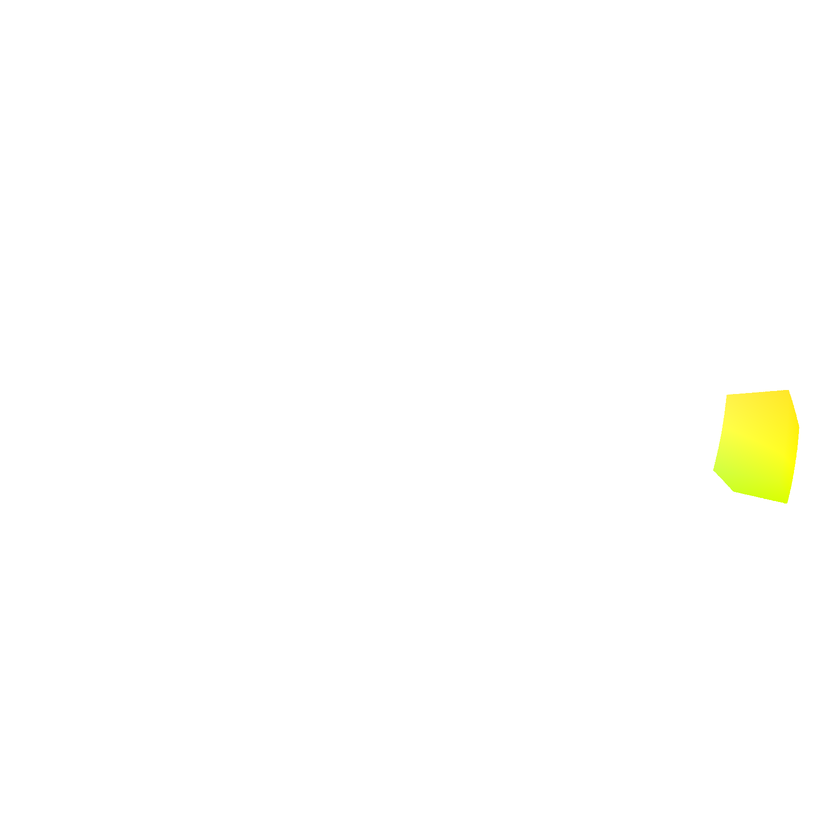}}\end{subfigure}
    \begin{subfigure}[b]{.19\linewidth}\fbox{\includegraphics[width=\textwidth]{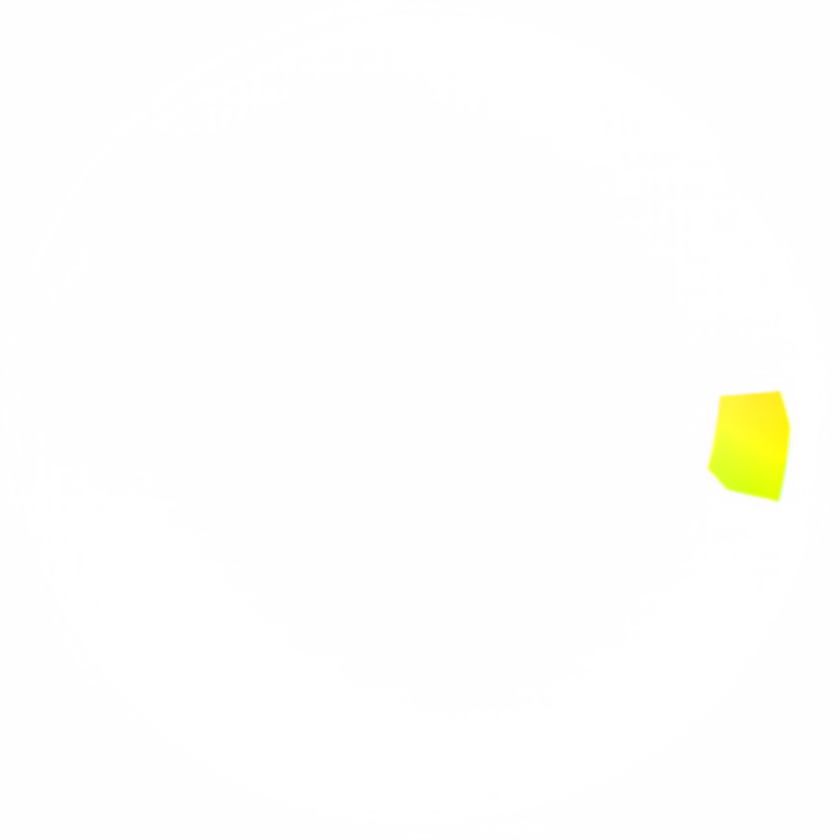}}\end{subfigure}
    \begin{subfigure}[b]{.19\linewidth}\fbox{\includegraphics[width=\textwidth]{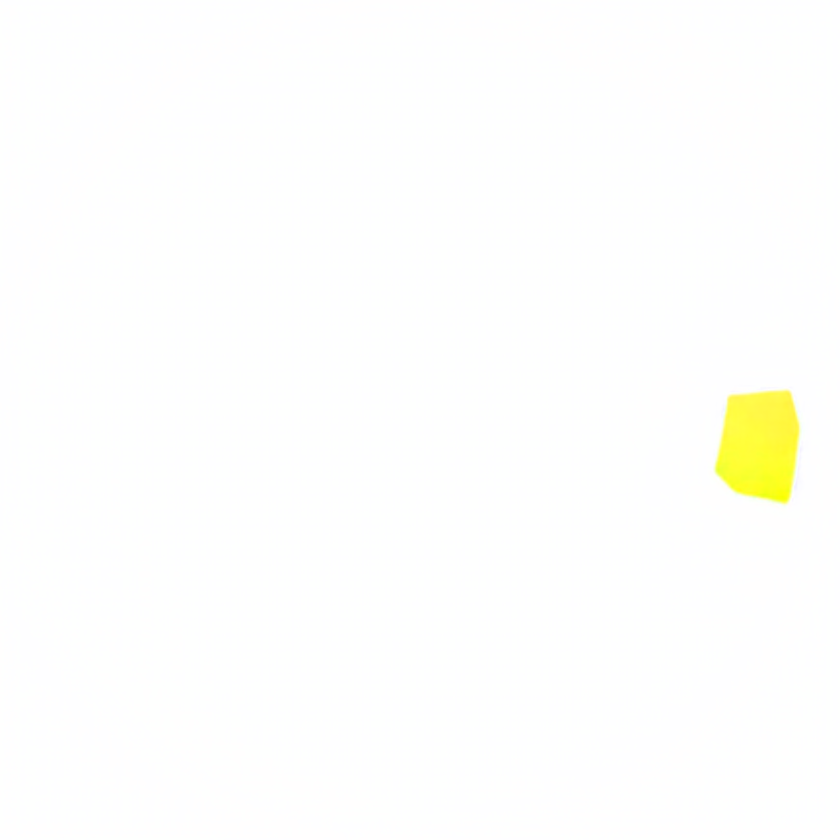}}\end{subfigure}
    \begin{subfigure}[b]{.19\linewidth}\fbox{\includegraphics[width=\textwidth]{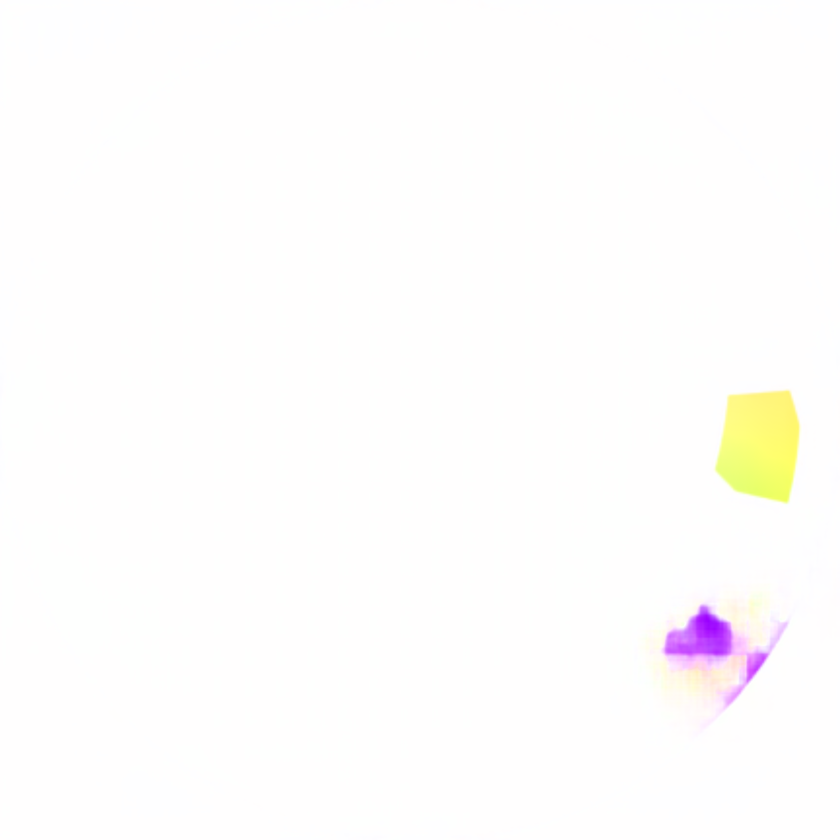}}\end{subfigure}

    \begin{sideways}
        \small linec-4, Frames 27\textrightarrow 28
    \end{sideways}
    \begin{subfigure}[b][0.195\textwidth]{.19\linewidth}\fbox{\includegraphics[width=\textwidth]{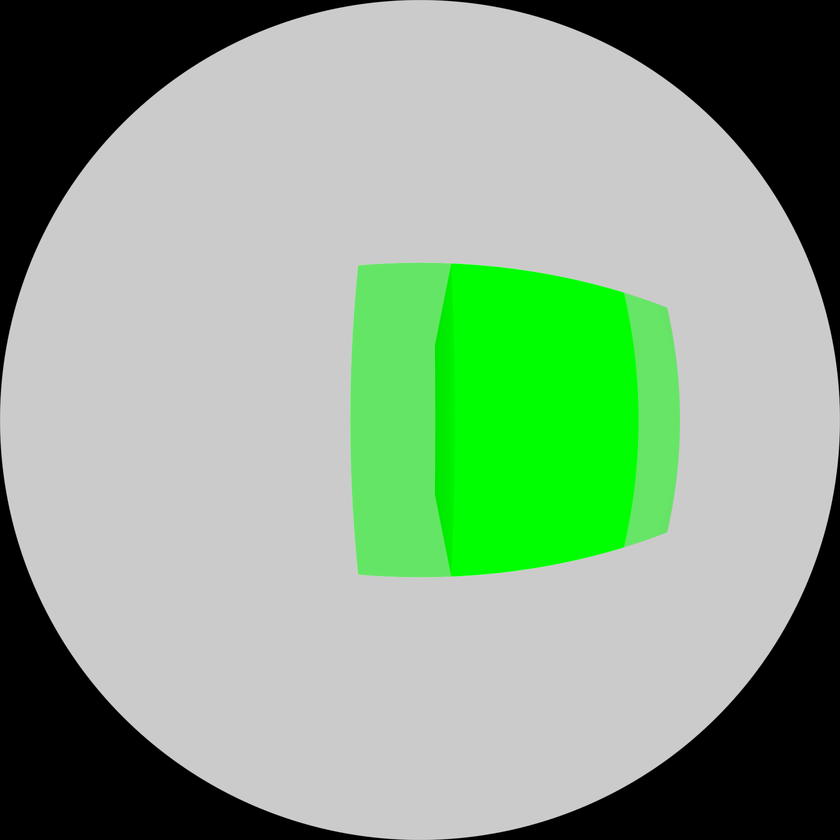}}\end{subfigure}
    \begin{subfigure}[b]{.19\linewidth}\fbox{\includegraphics[width=\textwidth]{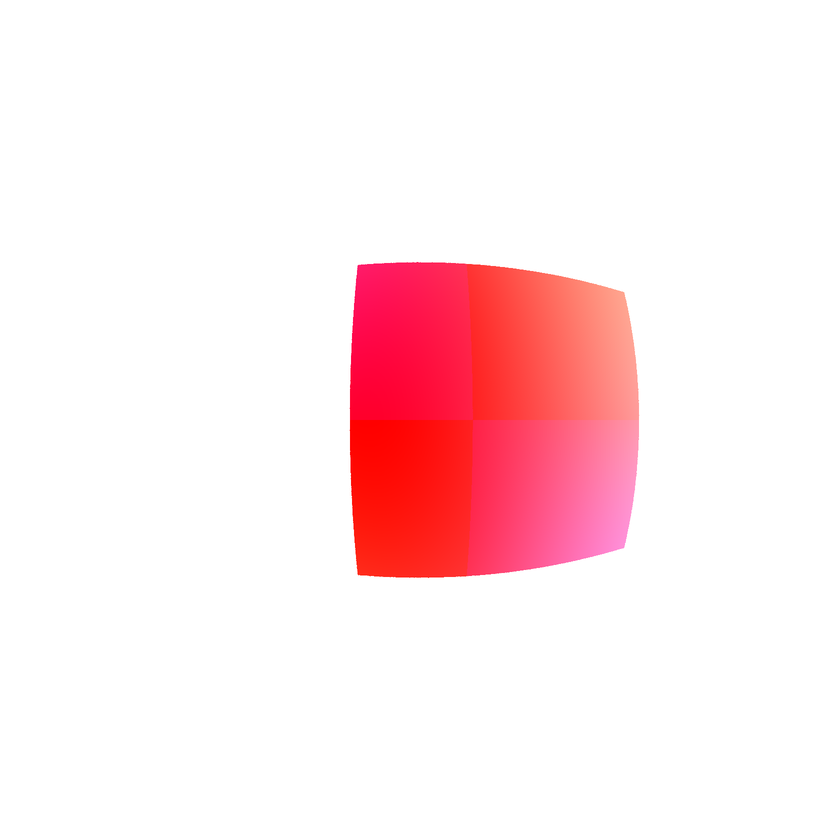}}\end{subfigure}
    \begin{subfigure}[b]{.19\linewidth}\fbox{\includegraphics[width=\textwidth]{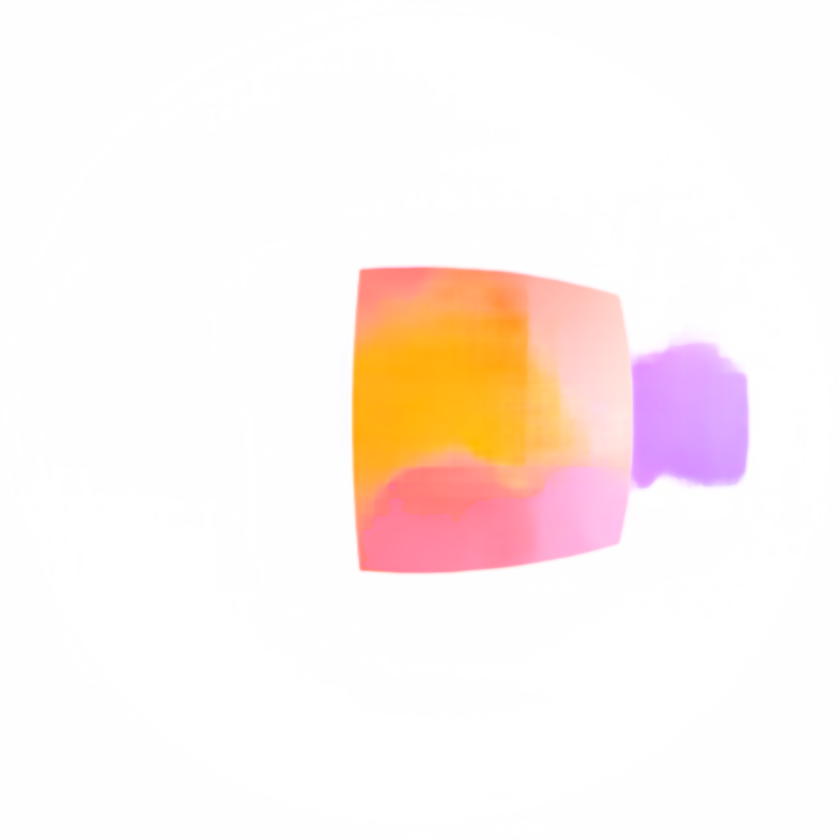}}\end{subfigure}
    \begin{subfigure}[b]{.19\linewidth}\fbox{\includegraphics[width=\textwidth]{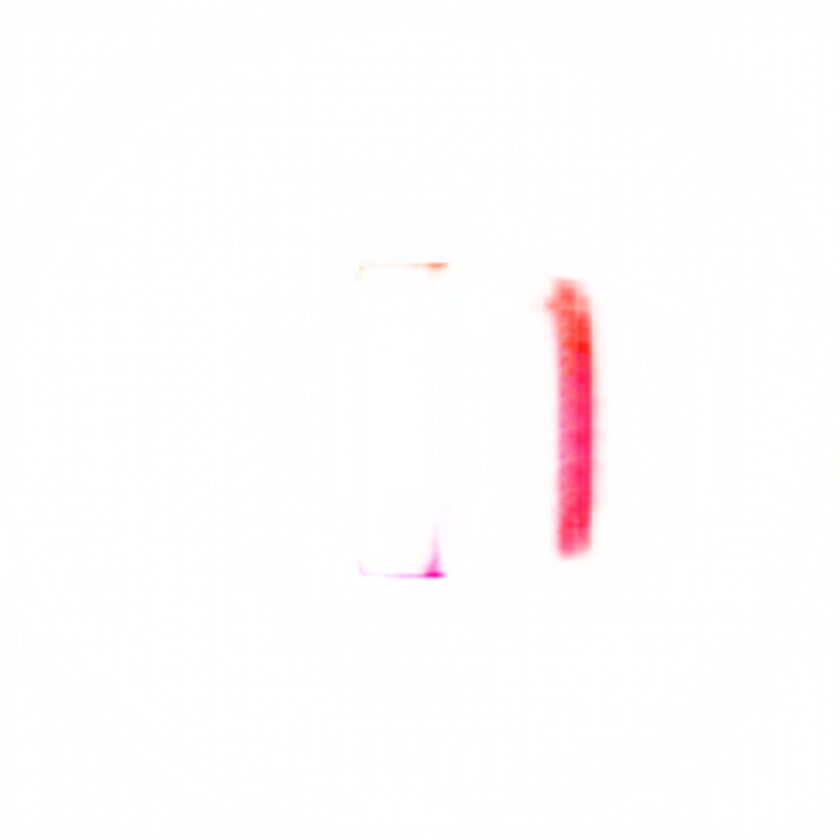}}\end{subfigure}
    \begin{subfigure}[b]{.19\linewidth}\fbox{\includegraphics[width=\textwidth]{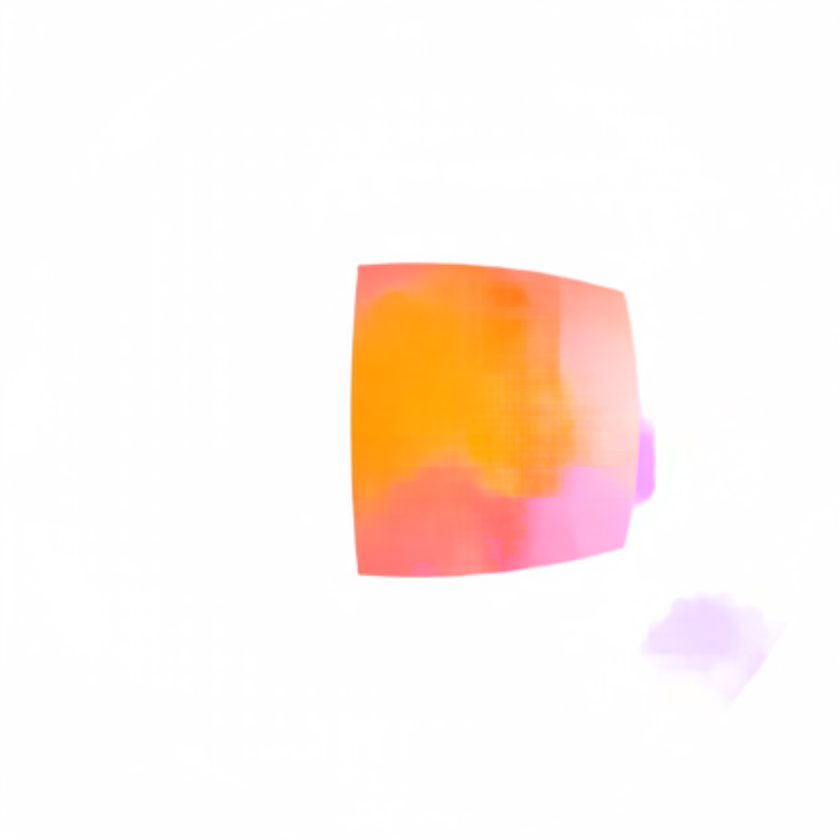}}\end{subfigure}

    \begin{sideways}
        \small linec-4, Frames 27\textrightarrow 28
    \end{sideways}
    \begin{subfigure}[b][0.195\textwidth]{.19\linewidth}\fbox{\includegraphics[width=\textwidth]{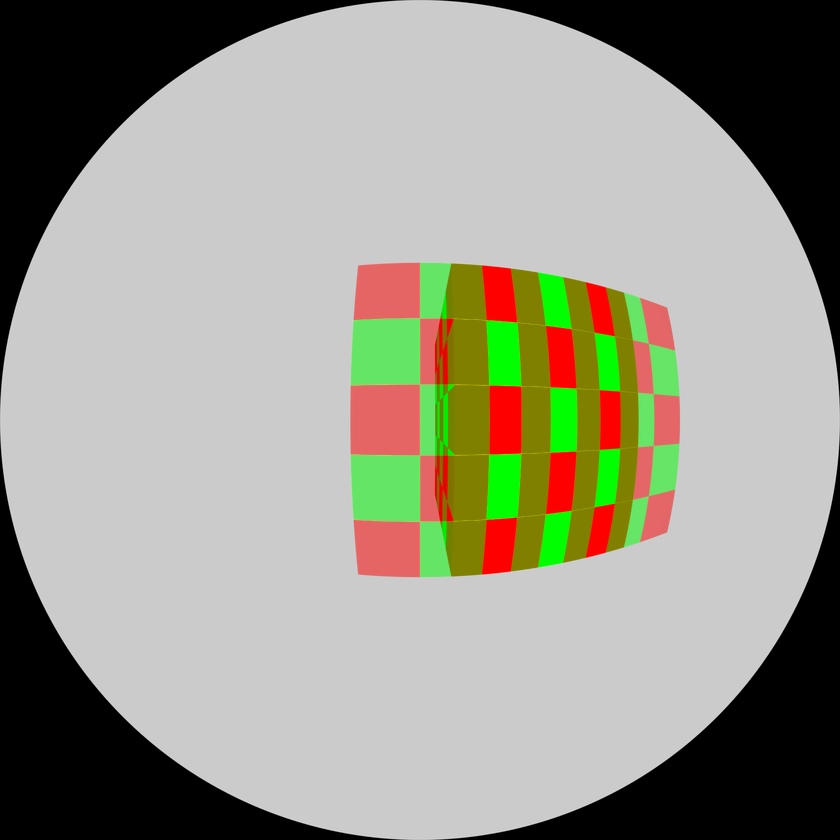}}\end{subfigure}
    \begin{subfigure}[b]{.19\linewidth}\fbox{\includegraphics[width=\textwidth]{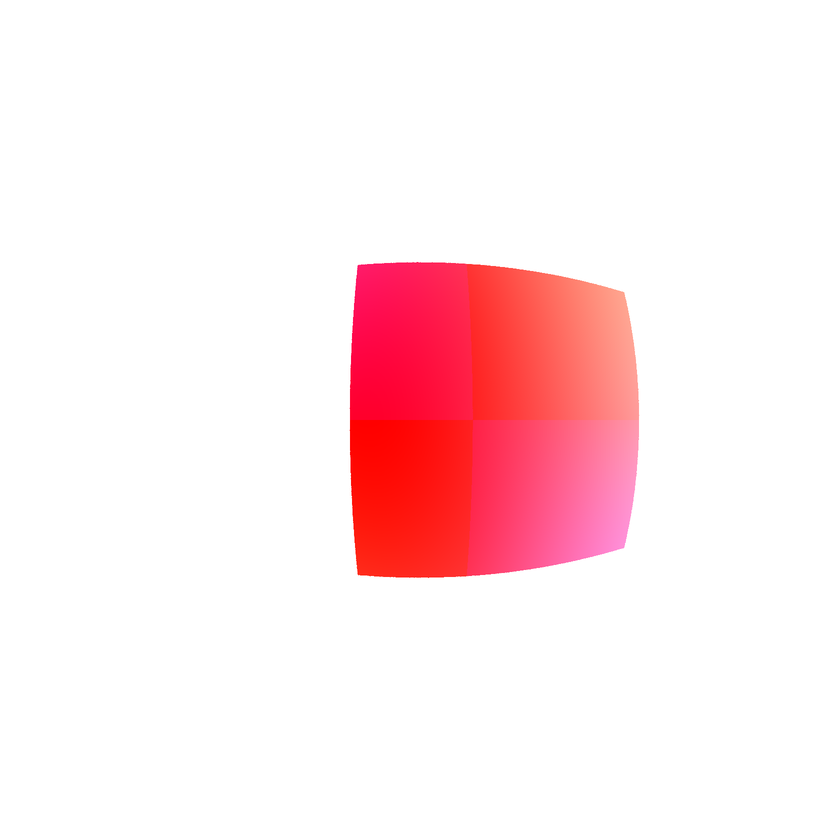}}\end{subfigure}
    \begin{subfigure}[b]{.19\linewidth}\fbox{\includegraphics[width=\textwidth]{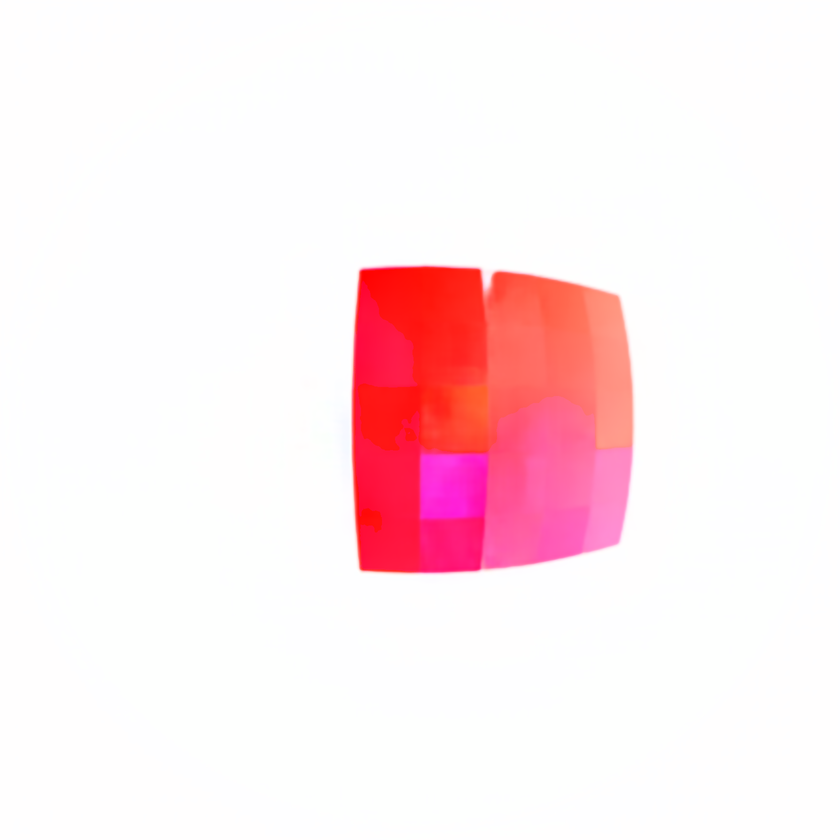}}\end{subfigure}
    \begin{subfigure}[b]{.19\linewidth}\fbox{\includegraphics[width=\textwidth]{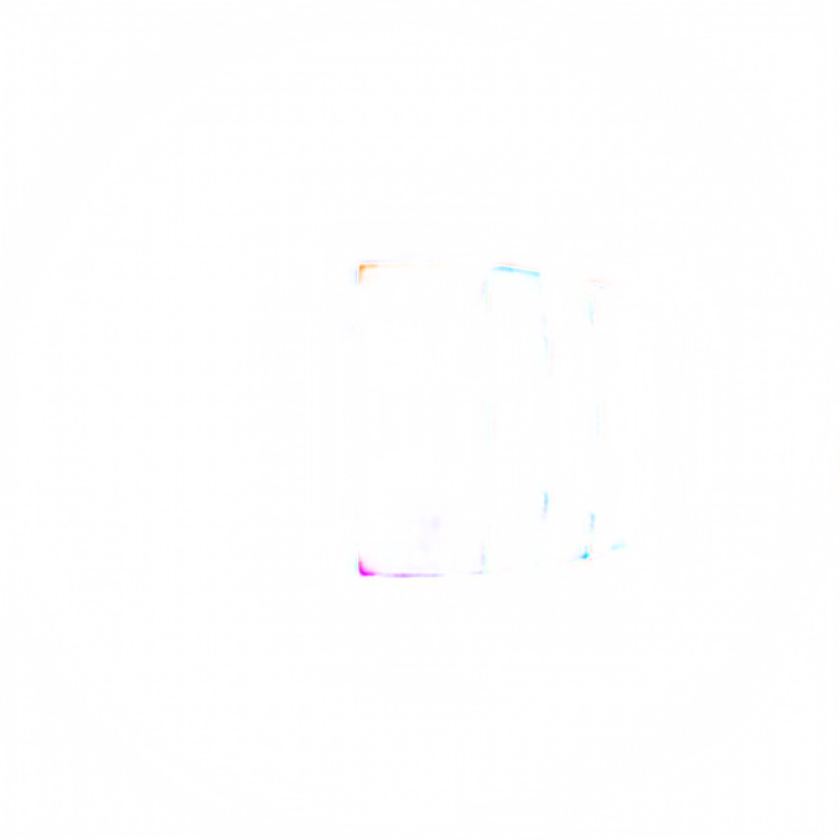}}\end{subfigure}
    \begin{subfigure}[b]{.19\linewidth}\fbox{\includegraphics[width=\textwidth]{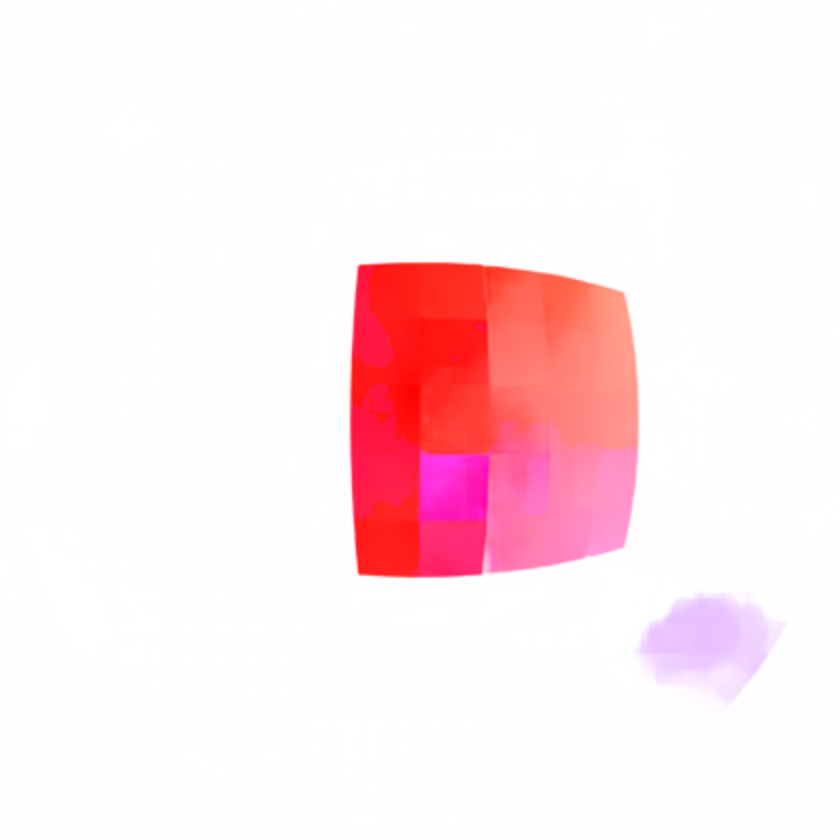}}\end{subfigure}

    \begin{sideways}
        \small spiral-4, Frames 2\textrightarrow 3
    \end{sideways}
    \begin{subfigure}[b][0.195\textwidth]{.19\linewidth}\fbox{\includegraphics[width=\textwidth]{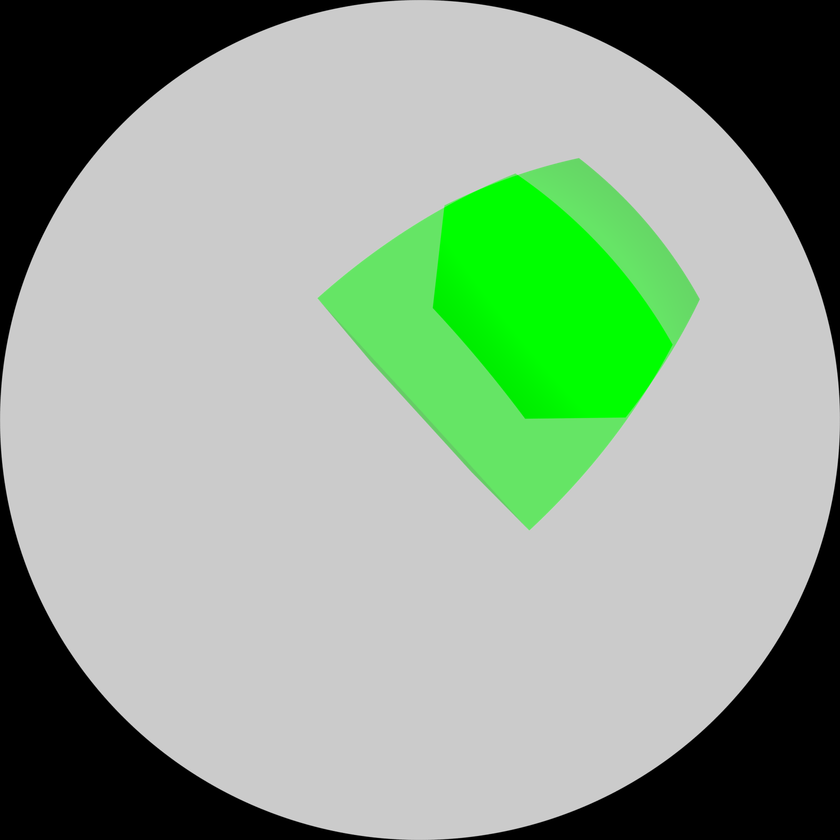}}\end{subfigure}
    \begin{subfigure}[b]{.19\linewidth}\fbox{\includegraphics[width=\textwidth]{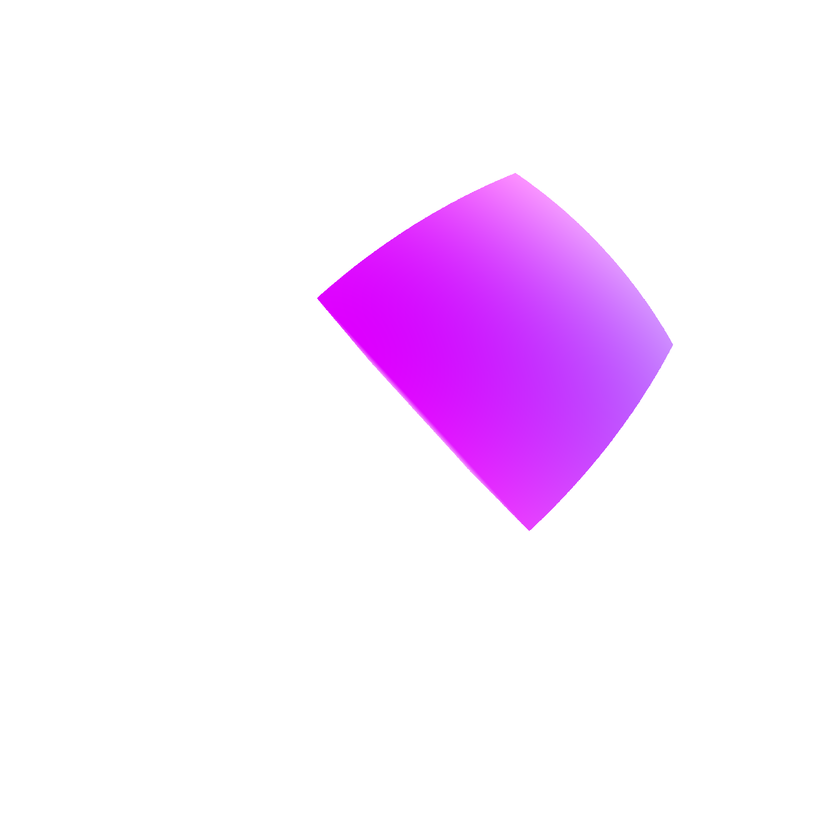}}\end{subfigure}
    \begin{subfigure}[b]{.19\linewidth}\fbox{\includegraphics[width=\textwidth]{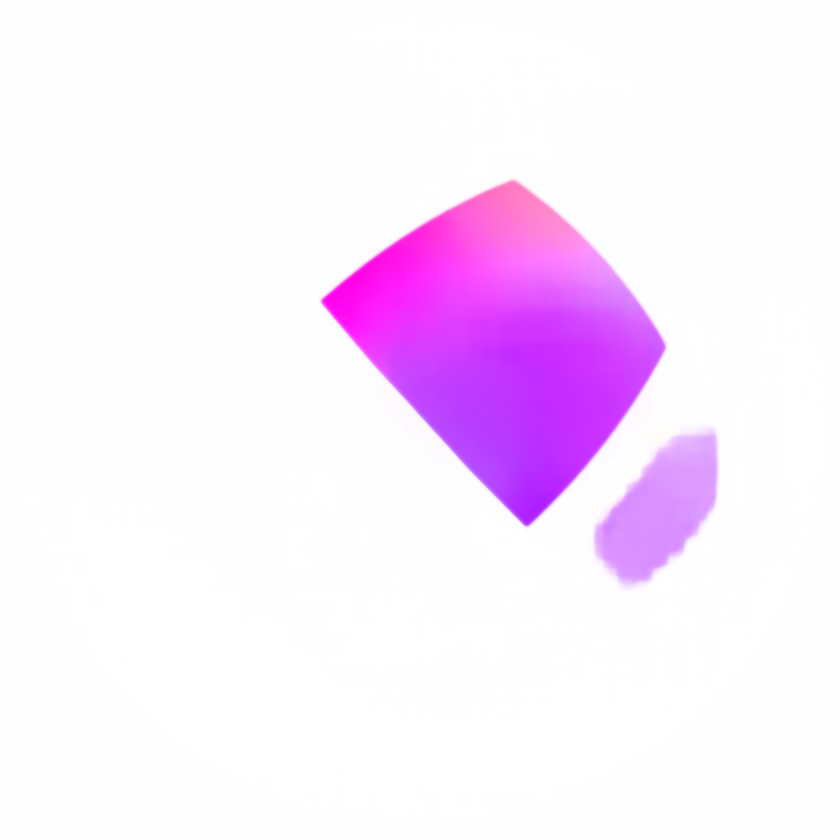}}\end{subfigure}
    \begin{subfigure}[b]{.19\linewidth}\fbox{\includegraphics[width=\textwidth]{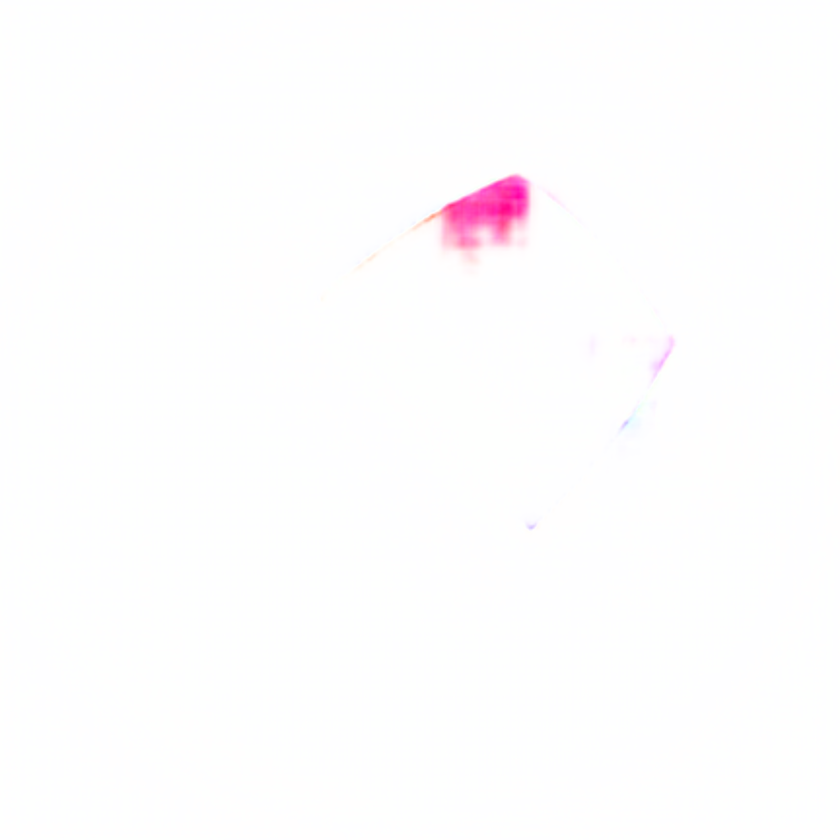}}\end{subfigure}
    \begin{subfigure}[b]{.19\linewidth}\fbox{\includegraphics[width=\textwidth]{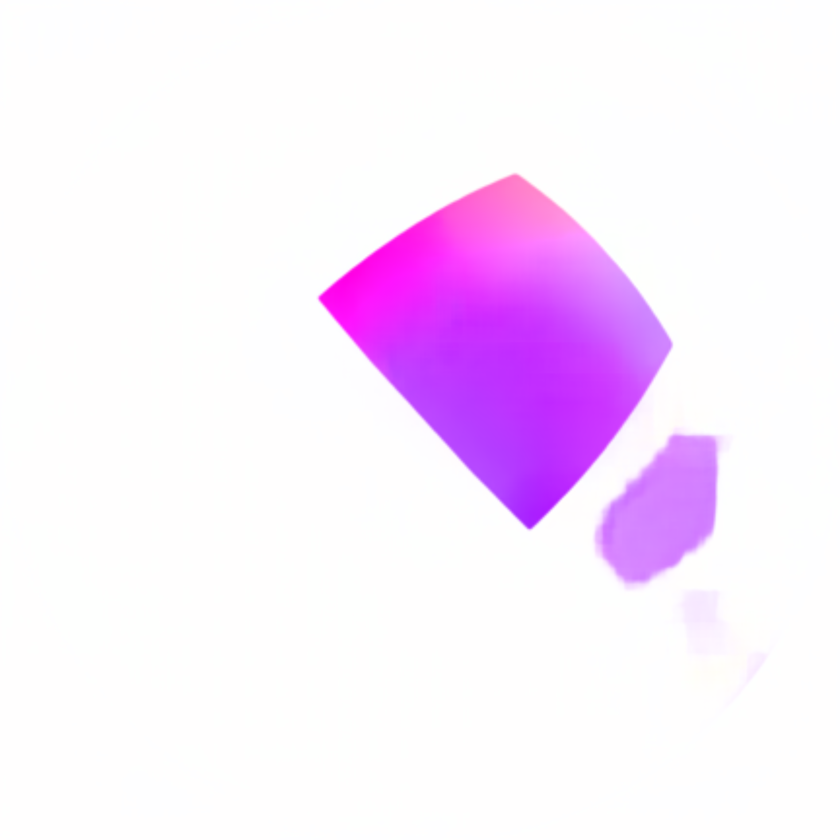}}\end{subfigure}
    \caption{Qualitative results on our ground truth. First column constitutes the semi-transparent superimposed input images. Second column shows the ground truth.
    Columns 3 to 5 contain various architectures, namely FlowNet2 (stacked networks), FlowNet2-SD for small displacements and FlowNet2-CSS-sd-ft (Refinement of network architecture).
    See text for details. We choose the examples of different architectures and various experiments to discover their limits intentionally.}\label{fig:qualit_res_all}
\end{figure*}

We formulate the selection of examples of the results in a way to determine different artifacts in our test data. The artifacts are induced by different characteristics, namely results of various architectures of FlowNet~2.0,
the way we synthetically generate ground truth and the camera geometry. A well-working example is shown in Fig.~\ref{fig:of_omni}.
Our investigations of different architectures based on FlowNet~2.0 induced by different experiments are shown in Fig.~\ref{fig:qualit_res_all}.

\setlength{\fboxsep}{0pt}

\subsection{Quantitative Evaluation of Results}


We evaluate the optical flow motion vector with respect to the synthetically generated ground truth by common error metrics,
the average angle error (AAE) and average end point error (AEPE) \cite{Baker2011}.
The angle error (AE) was first introduced in \cite{Fleet1990} and describes the angle between a flow vector \(\left(u,v\right)\) and a ground truth flow \(\left(\tilde u, \tilde v\right)\),
in 3D space between \(\left(u, v, 1\right)\) and \(\left(\tilde u, \tilde v, 1\right)\).
The AE can be computed by the inverse cosine of the dot product of the vectors divided through their magnitudes.
Averaged over all pixels of an image we get:
\begin{equation}
\mathrm{AE} = \frac{1}{N}\sum_{i=1}^N \cos^{-1}\left(\frac{1+u_i \cdot \tilde u_i + v_i \cdot \tilde v_i}{\sqrt{1+u_i^2+v_i^2}\sqrt{1+\tilde u_i^2+\tilde v_i^2}}\right)
\end{equation}
The relative measure of performance ensures to avoid the division by zero for zero flows.
By means of normalization the AE penalizes errors in large flows less than errors in small flows.
Current state-of-the-art flow algorithms present AAEs smaller than 2.0 degrees (equivalent to around 0.1 pixels) at Yosemite without clouds \cite{Barron1994} with effectively no outliers.

An absolute error for quantitative comparison of the results is the AEPE from \cite{Otte1994}.
The endpoint error (EPE) is described as the absolute magnitude of difference vectors of the flow vector and the corresponding ground truth vector.
Averaged over all pixels of an image we get:
\begin{equation}
\mathrm{EPE} = \frac{1}{N}\sum_{i=1}^N \sqrt{(u_i-\tilde u_i)^2 + (v_i^{}-\tilde v_i)^2}
\end{equation}

Evaluated results on AAE, AEPE and Flow errors are presented in Table~\ref{tab:aae_epe}.
For AAE and AEPE we compute the values of the optical flow from the variations of FlowNet~2.0 with respect to our generated synthetic motion vectors.
The mean values of the errors are calculated per experiment (variable number of images) and per architecture.
The best (i.e. lowest) values are highlighted in bold.

\begin{table*}[htb!]
\caption{Quantitative results of AAE, AEPE and Flow errors with different FlowNet~2.0 architectures.}
\label{tab:aae_epe}
\centering 
\begin{tabular}{llcclll}
\toprule
Exp. & Architecture & AAE [\si{\degree}] & AEPE [px] & Fl-bg [\si{\percent}] & Fl-fg [\si{\percent}] & Fl-all [\si{\percent}] \\
\midrule
linec-4 & FN2    & 1.92 & 2.73 & 0.91 & 55.42 & 3.38  \\
 & FN2-SD        & 1.72 & 2.73 & 0.06 & 44.66 & 2.53   \\
 & FN2-CSS-ft-sd & 4.54 & 2.13 & 1.22 & 37.59 & 3.43 \\
\cmidrule(l){2-7}
linec-2 & FN2    & 1.41 & 1.07 & 0.42 & 32.43 & 2.52  \\
 & FN2-SD        & 1.33 & 1.23 & 0.05 & 31.96 & 2.23  \\
 & FN2-CSS-ft-sd & 4.00 & 1.07 & 0.82 & 26.21 & 2.76  \\
\cmidrule(l){2-7}
linec-1 & FN2    & 1.08 & \textBF{0.43} & 0.11 & 22.85 & 1.96 \\
 & FN2-SD & \textBF{0.81} & 0.48 & \textBF{0.04} & 22.83 & \textBF{1.90} \\
 & FN2-CSS-ft-sd & 3.60 & 0.47 & 0.52 & \textBF{20.46} & 2.25 \\
\midrule
line-4 & FN2     & 0.65 & 0.08 & 0.12 & 40.86 & 0.34  \\
 & FN2-SD        & 0.16 & 0.05 & 0.05 & 59.03 & 0.42 \\
 & FN2-CSS-ft-sd & 3.02 & 0.09 & 0.29 & 33.40 & 0.49 \\
\cmidrule(l){2-7}
line-2 & FN2      & 0.63 & 0.04 & 0.11 & 28.97 & 0.26  \\
 & FN2-SD         & 0.12 & 0.02 & 0.03 & 26.93 & 0.20 \\
 & FN2-CSS-ft-sd  & 2.09 & 0.08 & 0.28 & 10.04 & 0.33 \\
\cmidrule(l){2-7}
line-1 & FN2      & 0.62 & 0.03 & 0.09 & 24.94 & 0.22  \\
 & FN2-SD         & {\textBF{0.10}} & \textBF{0.01} & \textBF{0.02} & \textBF{\phantom{0}4.75} & \textBF{0.04} \\
 & FN2-CSS-ft-sd  & 2.98 & 0.07 & 0.26 & \phantom{0}5.09 & 0.29 \\
\midrule
spiral-4 & FN2    & 1.42 & 1.71 & 0.49 & 81.46 & 2.05  \\
 & FN2-SD         & 0.93 & 2.14 & 0.10 & 94.29 & 1.81 \\
 & FN2-CSS-ft-sd  & 3.98 & 1.67 & 0.75 & 78.14 & 2.29 \\
\cmidrule(l){2-7}
spiral-2 & FN2    & 1.01 & 0.54 & 0.24 & 60.99 & 1.55 \\
 & FN2-SD         & 0.73 & 0.89 & 0.10 & 76.79 & 1.64 \\
 & FN2-CSS-ft-sd  & 3.45 & 0.57 & 0.50 & 48.43 & 1.71 \\
\cmidrule(l){2-7}
spiral-1 & FN2    & 0.86 & \textBF{0.24} & 0.14 & 37.77 & \textBF{1.05}  \\
 & FN2-SD         & \textBF{0.52} & 0.34 & \textBF{0.07} & 41.12 & 1.14 \\
 & FN2-CSS-ft-sd  & 3.22 & 0.29 & 0.36 & \textBF{23.80} & 1.15 \\
\bottomrule
\end{tabular}
\end{table*}

Taking into account the error metrics relative to the sum of pixel movement in terms of foreground and background, we follow the standard protocol from KITTI~2015 test set \cite{Menze2015CVPR,XuCVPR2017DCFlow}.
Due to this protocol we report percentage values for movements above 3 pixels of Flow-background (Fl-bg), Flow-foreground (Fl-fg) and Flow-all (Fl-all) to characterize the outliers of static background, dynamic foreground and the whole image.
Results are shown in Table~\ref{tab:aae_epe} in column 5.
On KITTI 2015 test dataset FlowNet~2.0 yields the best result in Fl-fg with \SI{8.75}{\percent}. 
The comparison to our own ground truth yields results of Fl-fg \SI{22.85}{\percent} at FlowNet2 architecture (per experiment values). 
Flownet2-SD architecture produces Fl-fg \SI{4.75}{\percent}. 
Flownet2-CSS-ft-sd leads to Fl-fg \SI{5.07}{\percent}. 
As long as our sequences contain a moving object with static background the Fl-bg values were close to zero while there is no motion in the background.
The Fl-fg errors of \SI{4.75}{\percent} and \SI{22.85}{\percent} are more representative and allows the comparison to the error at common benchmark datasets.

\noindent \textbf{We observe the following.}
(1) The average endpoint error of the sequence \textit{linec} decreases with lower velocity.
A plausible explanation is the higher proportion of foreground which is not evenly distributed because of the equidistant projection model of the sensor.
(2) We observe noisy effects in Fig.~\ref{fig:qualit_res_all} of results with the FN2-CSS-ft-sd architecture (column (e)) on the bottom-right location of the fish-eye image in all our experiments.
(3) Adding texture to the surface of the cube decreases the error of the foreground by about \SI{10}{\percent}. Our reference is the homogeneous (non-textured)
surface at linear motion through the center of the image (see Fig.~\ref{fig:qualit_res_all}, rows 3 and 4).
(4) As we expect, through the characteristics of the FlowNet2-SD architecture the determination of large displacements are difficult and leads to the highest values in Fl-fg
at especially \textit{spiral-1} (Table~\ref{tab:aae_epe}).
(5) At large displacements the stacked architecture of FlowNet2 generates a noisy smearing effect. This is shown in Fig.~\ref{fig:qualit_res_all} at column and row 3 and can be
reduced by adding texture to the object.

\section{Conclusion}
\label{sec:conclu}
Beside the intensity of the motion the texture of the object plays a significant role for large displacements.
In this paper we investigate the computation of optical flow by a state-of-the-art end-to-end learning approach, namely FlowNet~2.0 on omnidirectional image data.
For the evaluation of the results we synthetically generate cubes with a homogeneous and inhomogeneous surface and transform the motion vectors of the objects into 2-D image space using the generic camera model.
To step into the investigation of the determination of both, large and small displacements, either due to the characteristics of our camera model, or caused by an indeed high magnitude of the motion vector,
we apply the initial FlowNet2 architecture. The results have shown, that flow errors in foreground are higher than \SI{20}{\percent} and lead to noisy blurring effects even for large displacements.
To make the neural network more applicable for a wide range of displacements we see some prospective work on the training process. Is training on our simple geometric primitives qualitatively and quantitatively
sufficient to achieve competitive results on testing of common benchmark datasets?
Another application-driven work will be the motion segmentation based on FlowNet2's motion vectors on real-world data.

Finally, we note the improvement of the cubes3D dataset. Modelling the synthetic environment closer to the real world helps us to distinguish between foreground and background robustly.

\IEEEtriggeratref{14}

\bibliographystyle{IEEEtran}
\bibliography{IEEEabrv,lit}

\end{document}